\documentclass{article} %

\usepackage[utf8]{inputenc} %
\usepackage[T1]{fontenc}    %
\usepackage[pagebackref]{hyperref}       %
\usepackage{url}            %
\usepackage{booktabs}       %
\usepackage{amsfonts}       %
\usepackage{nicefrac}       %
\usepackage{microtype}      %
\usepackage{xcolor}         %
\usepackage{amsthm,amsmath,amssymb}
\usepackage[hmargin=3cm, vmargin=3cm]{geometry}

\usepackage{algorithm}
\usepackage{algorithmic}
\usepackage{multirow}
\usepackage{wrapfig}
\usepackage{caption}
\usepackage{subcaption}
\usepackage{natbib}

\usepackage[textsize=tiny]{todonotes}
\renewcommand*{\backrefalt}[4]{
  \ifcase #1 %
  No citations.%
  \or
  page~#2%
  \else
  pages~#2%
  \fi
}

\theoremstyle{remark}
\newtheorem*{remark}{Remark}

\theoremstyle{definition}
\newtheorem{definition}{Definition}[section]

\theoremstyle{plain}
\newtheorem{proposition}{Proposition}[section]

\definecolor{linkcolor}{RGB}{83,83,182}
\hypersetup{
    colorlinks=true,
    citecolor=linkcolor,
    linkcolor=linkcolor
}

\title{Detectability without Interference: Efficient Canary Design for Privacy Auditing in One Run}
\date{ }

\author{%
  Mathieu Dagréou$^{1}$\footnote{Corresponding author: \texttt{mathieu.dagreou@inria.fr}} , Valentin Dorseuil$^{2}$\footnote{Work done while visiting the Vector Institute.} , Mohammad Yaghini$^{3}$,\\
  Nicolas Papernot$^{3,4}$, Aurélien Bellet$^{1}$\\[0.5em]
  {\small $^1$PreMeDICaL team, Inria, Idesp, Inserm, Univ. de Montpellier}\\
  {\small $^2$DI ENS, École normale supérieure, Université PSL, CNRS}\\
  {\small $^3$University of Toronto \& Vector Institute}\\
  {\small $^4$Inria}
}

\newcommand{\bbE}{\mathbb{E}}

\newcommand{\bbP}{\mathbb{P}}

\newcommand{\bbR}{\mathbb{R}}

\DeclareMathOperator*{\argmin}{arg\,min}
\DeclareMathOperator*{\argmax}{arg\,max}

\newcommand{\clip}{\mathrm{Clip}}

\newcommand{\calA}{\mathcal{A}}

\newcommand{\calD}{\mathcal{D}}

\newcommand{\calI}{\mathcal{I}}

\newcommand{\calL}{\mathcal{L}}

\newcommand{\calO}{\mathcal{O}}

\newcommand{\calR}{\mathcal{R}}

\DeclareMathOperator{\Var}{Var}

\newcommand{\diff}{\mathrm{d}}

\usepackage[capitalise]{cleveref}
\crefname{figure}{Figure}{Figures}
\Crefname{figure}{Figure}{Figures}

\begin{document}

\maketitle

\begin{abstract}
  Privacy auditing aims to empirically assess privacy leakage in machine learning models using membership inference attacks (MIAs), and to derive lower bounds on differential privacy (DP) parameters. Standard privacy auditing inserts a sensitive sample (a canary) into many ($\gg 100$) randomized training runs, requiring the auditor to detect whether it was included in each run. In one-run auditing, many canaries are included in a single training run, avoiding the high computational cost of multi-run approaches. However, recent theory shows that canaries can interfere with one another, reducing per-canary MIA scores and thereby weakening the audit power. We propose IBIS, a framework that optimizes canaries to maximize their own score while minimizing their impact on others. IBIS combines influence-based greedy initialization with bilevel optimization, introducing a novel regularization term that encourages canaries to be orthogonal in embedding space as a proxy for reducing interference. We propose an efficient bilevel algorithm that incrementally updates a single model at each canary update, avoiding the need to retrain the model from scratch after every update. Experiments show that IBIS offers a better performance-cost trade-off than existing canary crafting approaches, improving MIA metrics while reducing computational cost by up to two orders of magnitude.
\end{abstract}

\section{Introduction}
Machine learning models are known to leak information about their training data \citep{carlini2019secret,Carlini2021,Haim2022,Nasr2025,barbero2025extracting}, motivating the need for privacy auditing. A popular approach to assess such leakage is through Membership Inference Attacks (MIAs) \citep{Shokri2017,Carlini2021,yeom2018privacy,Zarifzadeh2024,hayes2025strong}: given a data point $z$ and access to a trained model, an adversary attempts to infer whether $z$ was a \emph{member} of the training set, typically by constructing a confidence score derived from the model's outputs (e.g., based on the loss value at $z$). 
In the context of models trained with Differential Privacy (DP) \citep{Dwork2006}, one can further use MIA performance to obtain empirical lower bounds on the privacy parameters \citep{ding2018detecting,Jagielski2020,Nasr2021,Nasr2023,Cebere2025}. In this sense, privacy auditing complements DP accounting, which derives worst-case upper bounds on the same parameters.

A standard component of privacy auditing consists of inserting ``canary'' data points into the training set, intentionally designed to maximize detectable leakage, and measuring how accurately an adversary can infer whether a given canary was included in or excluded from the training data.
The usual \emph{multi-run auditing} approach randomizes the inclusion of a single canary at a time and repeats independent model training thousands of times to obtain statistically reliable estimates. This makes auditing prohibitively expensive for large-scale scenarios.
\emph{One-run auditing} \citep{Steinke2023,Mahloujifar2025,Even2026} instead considers a single training run in which $m$ canaries are independently included with probability $1/2$. Given the resulting model, an adversary then attempts to infer which canaries were included. This approach avoids the need to train a separate model for each randomized canary inclusion, substantially reducing the computational cost of privacy auditing.\looseness=-1

In the \emph{black-box} setting, where the adversary inserts sample canaries and observes only the final model's outputs, accurately detecting privacy leakage remains challenging \citep{Nasr2021,Nasr2023}. This challenge is particularly pronounced in the one-run paradigm, which generally yields weaker empirical estimates than multi-run approaches \citep{Keinan2025,Xiang2025}.
Several works \citep{Keinan2025,Xiang2025,Even2026} attribute this degradation to \emph{interference between canaries}: the inclusion or exclusion of a canary $z'$ can affect the MIA confidence score of another canary $z$, thereby degrading the adversary's ability to infer the membership of $z$. This interference is particularly strong when canaries are similar, as their effects on the model are correlated. In such cases, including $z'$ can reduce the model's sensitivity to $z$ and make its membership harder to detect.
Bridging the gap with multi-run auditing therefore requires canaries that are both \emph{detectable} in isolation and \emph{non-interfering} with one another: the MIA score of each canary should change significantly depending on its inclusion, but remain largely unaffected by the inclusion of other canaries.
Existing canary crafting methods designed for one-run auditing \citep{Boglioni2026meta_learn_canaries} do not explicitly promote \emph{non-interference}, limiting audit power; moreover, they are so expensive that they can effectively negate the computational advantage of the one-run paradigm.

\paragraph{Contributions.} We propose IBIS, a two-stage method for crafting canary sets for one-run auditing. IBIS first uses influence functions \citep{Koh2017} to greedily select training points that are highly detectable, as measured by their self-influence, while minimizing pairwise interference, as measured by their cross-influence. This provides a strong initialization for subsequent optimization. IBIS then refines the canaries through a bilevel optimization objective that jointly promotes detectability and orthogonality in representation space, using orthogonality as a tractable proxy for non-interference.

A key feature of our formulation is that it avoids evaluating canaries across the exponentially many possible subsets of included canaries. Instead, by promoting non-interference, we can compare the model trained with all canaries against a fixed reference model, requiring only a single model to be maintained throughout the optimization. We exploit this structure to develop an efficient bilevel algorithm that incrementally updates the model after each canary update, rather than retraining it from scratch at every outer step. This substantially reduces the cost of canary crafting while preserving strong attack performance.

We evaluate IBIS on WideResNet models trained on CIFAR-10 and CIFAR-100, and on a multilayer perceptron trained on MNIST. Our results show that IBIS effectively reduces interference between canaries, leading to higher privacy leakage estimates than prior canary crafting methods for one-run auditing. At the same time, IBIS is substantially more computationally efficient, reducing the cost of canary crafting by up to two orders of magnitude. This efficiency allows the improved audit performance to retain the computational advantage of one-run auditing over multi-run approaches.

\paragraph{Related work on canary crafting.} Heuristic canary crafting methods are routinely used, such as flipping the label of a sample or using a blank image \citep{Nasr2023}. The seminal work of \citet{Jagielski2020} introduced the ClipBKD technique, which optimizes the canary for multi-run auditing by aligning it with the lowest-variance direction of the training data covariance.
Several follow-up works also operate primarily in the multi-run setting. \citet{Lu2022} use influence functions to design canaries that maximize their effect on model parameters, but instantiate them only for logistic regression. More recently, \citet{Yaghini2026optifluence} leverage influence functions and bilevel optimization, but optimize a single canary rather than a set. Closest to us, \citet{Boglioni2026meta_learn_canaries} target the one-run regime, but do not promote non-interference and require retraining the model at every canary update. \citet{Yaghini2026optifluence} and \citet{Boglioni2026meta_learn_canaries} both rely on unrolling the training loop, as discussed in \cref{sec:frame_bilevel}. In the federated setting, \citet{Maddock2023} craft a canary by penalizing the inner products between its gradient and the other clients' updates, while keeping its gradient norm large.
A related line of work considers a different threat model where the auditor injects gradients rather than data points, enabling tight privacy audits in certain settings \citep{Cebere2025}.\looseness=-1

\section{Background}

\subsection{Membership inference and differential privacy}
Membership Inference Attacks (MIAs) \citep{Shokri2017,Carlini2022mia} are privacy attacks in which an adversary attempts to infer whether a given data point $z$ was part of the training set of a model $\calA(D)$ trained on dataset $D$. The adversary computes a score $s(\calA(D), z)$, for instance based on predicted probabilities or loss values, and thresholds it to produce a binary guess. MIAs are classically formulated as a hypothesis test where the null hypothesis is that $z \notin D$, yielding standard performance metrics such as the Area Under the Curve (AUC) or the True Positive Rate (TPR) at low False Positive Rate (FPR) \citep{Carlini2022mia}: the higher these metrics, the stronger the MIA.

Differential Privacy (DP) is the standard framework to quantify the privacy guarantees of randomized algorithms \citep{Dwork2014f&t,Dwork2006}. It ensures that the algorithm's output distribution does not change significantly when a single input data point is added or removed.
\begin{definition}
A randomized algorithm $\calA:\calD \to \calO$ satisfies $(\epsilon,\delta)$-DP if, for any datasets $D,D'\in\calD$ s.t. $D'=D\cup\{z\}$ or $D=D'\cup\{z\}$, and for any set $O\subseteq\calO$,
$
    \bbP[\calA(D)\in O]
    \leq
    e^\epsilon \bbP[\calA(D')\in O] + \delta.
$
\end{definition}
\vspace{-.2cm}
The parameter $\epsilon$ controls the strength of the guarantee, with smaller values providing stronger privacy, while $\delta$ should be a small constant that allows a negligible probability of failure. At scale, DP training is usually performed with DP-SGD \citep{Song2013,Bassily2014,Abadi2016DeepDP}, which clips individual gradients and adds Gaussian noise to the update direction.

Importantly, DP also limits the success of any MIA \citep{Nasr2021}. MIA-based privacy auditing can thus provide empirical lower bounds on DP guarantees, in addition to measuring privacy leakage for models without formal privacy guarantees. We next recall the basics of one-run privacy auditing.

\subsection{Privacy auditing in one run}
\label{ssec:auditing}

\begin{wrapfigure}{r}{0.5\linewidth}
  \vspace{-0.83cm}
  \begin{minipage}{\linewidth}
    \begin{algorithm}[H]
      \begin{algorithmic}[1]
        \STATE {\bfseries Input:} Dataset $D$, audited algorithm $\calA$, score function $s$, canaries $C =\{z_1,\dots,z_m\}$
        \STATE Draw $A\sim\mathrm{Bernoulli}(1/2)^{\otimes m}$.
        \STATE Compute $\theta = \calA(D\cup\{z_i : A_i = 1\})$
        \FOR{$i = 1,\dots, m$}
          \STATE Compute the score $\hat{s}_i = s(\theta, z_i)$\label{line:alg_score}
        \ENDFOR
          \STATE $\mathrm{MIA\_perf} = \mathrm{TPR}_{@\alpha\text{-FPR}}(\hat{s}, A)$
          \STATE $\mathrm{DP\_lower\_bound} = \hat\epsilon(\hat{s}, A)$
        \STATE {\bfseries Output:} $\mathrm{MIA\_perf}, \mathrm{DP\_lower\_bound}$
      \end{algorithmic}
      \caption{Auditing in one run \citep{Steinke2023,Even2026}}
      \label{alg:auditing}
    \end{algorithm}
  \end{minipage}
  \vspace{-.4cm}
\end{wrapfigure}
In one-run auditing \citep{Steinke2023,Even2026}, the auditor has a set $C = \{z_1,\dots,z_m\}$ of canaries. A selection $S = \{z_i : A_i = 1\}$ is built by drawing $A_i\sim\mathrm{Bernoulli}(1/2)$ independently for each canary. The model is then trained on $D\cup S$, yielding model parameters $\theta$. A MIA score $s(\theta, z)$ is computed for each canary $z\in C$. This function should be such that $s(\theta, z)$ is high when $z\in S$ and low otherwise. In our experiments, we use the loss-based score $s(\theta, z) = -\ell(\theta, z)$ where $\ell$ is the training loss (e.g. the cross-entropy loss), which is a simple and commonly used choice. We do not use stronger attacks such as LiRA \citep{Carlini2022mia} or RMIA \citep{Zarifzadeh2024} since they rely on shadow models, which is not relevant in the one-run auditing context. The procedure is summarized in \cref{alg:auditing}.

These scores alongside the ground-truth memberships can then be used for privacy auditing.
One can compute standard attack performance metrics, such as TPR at low FPR \citep{Carlini2022mia}:
$$
  \mathrm{TPR}_{@\alpha}(\hat s, A) = \max_{\tau}\left\{ \frac{\sum_i \mathbf{1}[\hat{s}_i \geq \tau]\, A_i}{\sum_i A_i} \;:\; \frac{\sum_i \mathbf{1}[\hat{s}_i \geq \tau]\,(1-A_i)}{\sum_i (1-A_i)} \leq \alpha \right\}.
$$
For DP auditing, following \citet{Steinke2023}, we can construct a lower bound of the privacy parameter $\epsilon$. To do so, the $k_+$ highest scores are guessed to have been included in the training set, while the $k_-$ lowest scores are guessed to have been excluded. From these statistics, the auditor can either directly estimate a lower bound on $\epsilon$ \citep{Steinke2023}, or estimate the Gaussian Differential Privacy (GDP) parameter \citep{Dong2019} for a tighter bound \citep{Nasr2023,Mahloujifar2025}.\looseness=-1

Audit quality depends heavily on the choice of canaries. Indeed, the inclusion of a canary should strongly affect its own score while minimally affecting others. Influence functions provide a proxy for these properties; we will use them to select a good initial set of canaries from the dataset.

\subsection{Influence functions}\label{ssec:influence_functions}
Influence functions aim at measuring the effect of upweighting a data sample on a model. They were first introduced in robust statistics \citep{Hampel1974} and then applied in machine learning for various purposes such as explainability \citep{Koh2017}, or to study the memorization phenomenon in neural networks \citep{Feldman2020,Zhang2023counterfactual_memorization,Meeus2025influence} thanks to the notion of cross and self-influence. Recently, the computational cost of influence functions has been reduced so that they can be used at scale \citep{Bae2022,Grosse2023}.
Consider a dataset $D= \{z_1,\dots,z_n\}$, a loss function $\ell$ and the solution of the Empirical Risk Minimization (ERM) problem $\theta^*(D) = \argmin_\theta \calL(\theta, D)$, where $\calL(\theta, D) \triangleq \frac1n \sum_{i=1}^n\ell(\theta, z_i)$. For $\alpha>0$ and a data point $z$, we denote $\theta^*_{z}(\alpha;D)$ the model obtained by solving $\min_\theta \frac1n \sum_{i=1}^n \ell(\theta, z_i) + \alpha \ell(\theta, z)$.
Since $\theta^*(D) = \theta^*_z(0;D)$, the difference $\theta^*_{z}(\alpha;D)-\theta^*(D)$ can be approximated by the Jacobian~$\alpha\partial_\alpha \theta^*_{z}(0;D)$. Under strong convexity of the loss function $\ell$, the implicit function theorem yields the \emph{influence function} of $z$: $\partial_\alpha \theta^*_{z}(0;D) = -\big[\nabla^2\calL(\theta^*(D), D)\big]^{-1}\nabla \ell(\theta^*(D), z)$.
Then, for a measurement function $f$, substituting the influence function into a first-order Taylor expansion yields $f(\theta^*_{z}(\alpha;D)) \approx f(\theta^*(D)) + \alpha \nabla f(\theta^*(D))^\top \partial_\alpha \theta^*_{z}(0;D)$.

\section{Canary selection by influence optimization}\label{sec:preselection}

The work of \citet{Yaghini2026optifluence} highlights the value of influence-based initialization for gradient-based canary crafting in the multi-run setting. In the one-run setting, however, multiple canaries coexist in the same training run, making their mutual interference a key consideration. We therefore use both self- and cross-influence to greedily select a \emph{set} of dataset points that are individually detectable while having low mutual interference, and use this set to initialize our subsequent bilevel optimization.\looseness=-1

\subsection{Cross- and self-influence on MIA scores}\label{ssec:cross_self_influence}

Consider any MIA score function $s$ such that $s(\theta^*(D), z)$ is higher when $z \in D$. To quantify detectability and interference, we track how the score of $z$ changes when a point $z'$ is removed from $D$. Applying the first-order approximation from \cref{ssec:influence_functions} with $\alpha = -1/|D|$ gives $s(\theta^*(D), z) - s(\theta^*(D\setminus\{z'\}), z) \approx \frac{1}{|D|}\calI_s(z, z')$, with $\calI_s(z, z')$ defined in \cref{def:cross_self_influence}.

\begin{definition}[Self- and cross-influence]\label{def:cross_self_influence}
  The \emph{influence} of $z'$ on the score of $z$ is $\calI_s(z, z') \triangleq \nabla_\theta s(\theta^*(D), z)^\top \partial_\alpha\theta^*_{z'}(0; D)$. When $z' = z$, $\calI_s(z,z)$ is the \emph{self-influence} of $z$; when $z' \neq z$, $\calI_s(z,z')$ is the \emph{cross-influence} of $z'$ on $z$.
\end{definition}
\begin{remark}
  Under strongly-convex ERM and the loss-based score $s(\theta,z) = -\ell(\theta,z)$, both reduce to $\calI_s(z,z') = \nabla\ell(\theta^*(D),z)^\top[\nabla^2\calL(\theta^*(D),D)]^{-1}\nabla\ell(\theta^*(D),z')$. Self-influence $\calI_s(z,z) \geq 0$ recovers the intuition that removing $z$ increases its loss. A positive cross-influence $\calI_s(z,z') > 0$ means that removing $z'$ also increases the loss of $z$, which creates interference.
\end{remark}

In the one-run auditing setting, a large $\calI_s(z_i, z_j)$ means that the inclusion or exclusion of $z_j$ changes the score of $z_i$ in line~\ref{line:alg_score} of Alg.~\ref{alg:auditing}, making its membership harder to infer. This motivates selecting canaries with low pairwise cross-influence.

\subsection{Greedy search of canary set}

In this section, we introduce the first step of our canary-crafting approach: selecting initial canary candidates from the training set $D$. These candidates initialize the non-convex gradient-based optimization procedure of \cref{sec:refine}.

\begin{wrapfigure}{r}{0.53\linewidth}
  \vspace{-.9cm}
  \begin{minipage}{\linewidth}
    \begin{algorithm}[H]
        \begin{algorithmic}[1]
          \STATE {\bfseries Input:} Dataset $D$, number of canaries $m$, number of preselected canaries $p$
          \STATE Let $\overline{D} \subset D$ be the $p$ samples with highest self-influence.
          \STATE Set $C_1 = \{\argmax_{z\in \overline{D}}\calI_s(z,z)\}$
          \FOR{$k = 1,\dots, m-1$}
          \STATE Set $z_{k+1} = \argmax_{z\in  \overline{D}\setminus C_k}\frac{\calI_s(z, z)}{\max_{z'\in C_{k}} \calI_s(z, z')}$
          \STATE Update $C_{k+1} = C_k \cup \{z_{k+1}\}$
          \ENDFOR
          \STATE {\bfseries Output:} $C_m$
        \end{algorithmic}
        \caption{Greedy canary selection}
      \label{alg:canary_selection}
    \end{algorithm}
  \end{minipage}
  \vspace{-0.5cm}
\end{wrapfigure}

Our selection strategy leverages influence functions to identify a set $C~=~\{z_1, \dots, z_m\} \subset D$ whose elements exhibit high detectability while minimizing mutual interference. Specifically, we require that the inclusion of each canary $z_i \in C$ can be reliably inferred from its MIA score $s(\theta, z_i)$, and that this score remains largely insensitive to the presence or absence of any other canary $z_j \in C$, with $z_j \neq z_i$.

We use self-influence as a proxy for detectability and cross-influence as a proxy for mutual interference. Consequently, we seek a set $C$ in which each canary has high self-influence and low cross-influence with the other selected canaries. We only control cross-influence within $C$, since the remaining points of $D$ are included in every training run. Their effect on a canary's score is therefore constant across runs: they may shift the score, but do not cause it to fluctuate.
Formally, for canary $z_i\in C$, $\calI_s(z_i,z_i)$ measures the change in its score induced by its own inclusion, whereas $\calI_s(z_i,z_j)$ measures the change induced by another canary $z_j\in C\setminus\{z_i\}$. The ratio
$\frac{\calI_s(z_i,z_i)}{\max_{z_j\in C\setminus\{z_i\}}\calI_s(z_i, z_j)}$
therefore captures the canary's own signal relative to its strongest potential source of interference. 
This leads to the following combinatorial optimization problem:\looseness=-1
\begin{align}\label{eq:opt_influence_pb}
  \max_{C\subset D, \lvert C\rvert = m} \Big\{f(C) = \sum_{z_i\in C}\frac{\calI_s(z_i,z_i)}{\max_{z_j\in C\setminus\{z_i\}}\calI_s(z_i, z_j)}\Big\}.
\end{align}
Solving \eqref{eq:opt_influence_pb} exactly requires evaluating $\binom{|D|}{m}$ candidate subsets, which is computationally infeasible. Moreover, since $f$ is not submodular, no efficient algorithm with approximation guarantees is known. We therefore adopt a greedy heuristic \citep{Papadimitriou1982}, summarized in \cref{alg:canary_selection}.

The procedure begins with a preselection step: we construct a subset $\overline{D}\subset D$ containing the $p$ samples with highest self-influence. This reduces the number of required cross-influence evaluations. We then build $C$ iteratively. Given a partial set $C_k$ of size $k$, the next canary is chosen by maximizing, over $z\in \overline{D}\setminus C_k$, the ratio~$\frac{\calI_s(z, z)}{\max_{z'\in C_{k}} \calI_s(z, z')}$. This requires computing and storing all pairwise self- and cross-influences within $\overline{D}$, resulting in a $p \times p$ score matrix. The computation of the influence functions, which involves inverse Hessian-vector products, can be made more efficient in large scale setting by leveraging K-FAC approximation of the Hessian \citep{Martens2015,Bae2022}. \looseness=-1

As we will see experimentally, this selection provides a good initialization for the gradient-based refinement step introduced next, which searches for canaries beyond the training set.

\section{Gradient-based refinement of canaries}
\label{sec:refine}

In this section, we design a bilevel objective function to craft canaries that are highly detectable and exhibit low mutual interference. This objective is optimized through a gradient-based method initialized with the canary set we obtain after the greedy search (\cref{alg:canary_selection}).

\subsection{Warm-up: canary interference in least squares regression}\label{ssec:warmup}
 To build intuition, we first consider the simplest setting: least squares regression with two canaries.
 It has the advantage of admitting a closed-form solution. Let $D=\{z_i=(x_i, y_i)\}_{1\leq i\leq n}$ denote the training set, where $x_i\in\bbR^d$ represents input features and $y_i\in\bbR$ the corresponding label. Denoting $X = (x_1,\dots,x_n)^\top\in\bbR^{n\times d}$ the design matrix and $y = (y_1,\dots,y_n)^\top\in\bbR^n$ the target vector, we seek $\theta^*(X, y) \in \argmin_{\theta\in\bbR^d} \frac12\|X\theta - y\|^2$.
When $n\geq d$ and $X$ has full rank, the solution is unique and given by $\theta^*(X, y) = (X^\top X)^{-1}X^\top y$.

Consider now two canaries $z_{c,1} = (x_{c,1}, y_{c,1})$ and $z_{c,2}= (x_{c,2}, y_{c,2})$. We define 
$
  \tilde X_i = \begin{bmatrix}
    X \\
    x_{c, i}
  \end{bmatrix}$, and  $\tilde y_i = \begin{bmatrix}
        y \\
        y_{c,i}
        \end{bmatrix}$.
Let $\ell(\theta,z_i)=\frac{1}{2}(y_i - \theta^\top x_i)^2$ denote the least square loss. We are interested in quantifying how the inclusion of $z_{c,1}$ in the training set affects the loss evaluated at $z_{c,2}$, i.e., how $\ell(\theta^*(\tilde X_1, \tilde y_1), z_{c,2})$ differs from $\ell(\theta^*(X, y), z_{c,2})$. The following proposition characterizes this effect.

\begin{proposition}\label{prop:gap_canary_ols}
    Assume that $n\geq d$ and that the matrix $X$ has rank $d$. Then we have
    \begin{align}\label{eq:gap_canary_ols}
        \ell(\theta^*(\tilde X_1, \tilde y_1), z_{c,2}) - \ell(\theta^*(X, y), z_{c,2}) = \gamma\; x_{c,1}^\top K^{-1} x_{c,2},
    \end{align}
    where $K = X^\top X$ and $\gamma$ is a scalar that depends on the two canaries and on $\theta^*(X, y)$. Its closed form is given in \cref{app:proof_gap_canary_ols}.
\end{proposition}
We provide the proof in \cref{app:proof_gap_canary_ols}. Equation \eqref{eq:gap_canary_ols} vanishes as soon as $x_{c,1}^\top K^{-1} x_{c,2} = 0$, that is, as soon as the two canaries are \emph{$K^{-1}$-orthogonal}. This observation motivates incorporating an orthogonality constraint into our bilevel objective. While this derivation is specific to least squares, it transfers to deep models through their last layer, which fits a linear map on the learned representations. Applying \cref{prop:gap_canary_ols} at that layer replaces the inputs by the embeddings $\varphi_{\theta^*}(x)$ computed by the preceding layers, and $K$ by their Gram matrix. We therefore promote orthogonality in the representation space induced by the model, rather than in raw input space.

\subsection{Canary optimization as a bilevel optimization problem}
\label{sec:frame_bilevel}

Unlike \cref{sec:preselection}, where canaries are selected from a fixed dataset $D$, we now treat the canary samples themselves as optimization variables and formulate a bilevel objective amenable to gradient-based optimization.
Throughout this section, we slightly abuse notation and write $\theta^* \triangleq \theta^*(D)$ and $\theta_S^* \triangleq \theta^*(D\cup S)$ for any set $S$, where $\theta^*(D') \in \argmin_\theta \calL(\theta, D')$.

The canaries obtained via \cref{alg:canary_selection} already exhibit low pairwise cross-influence by construction, implying that for any canary $z\in C$ and subset $S\subset C$, the score $s(\theta_S^*, z)$ is approximately determined only by whether $z\in S$. Under this approximation, it suffices to consider the full inclusion case $S=C$, yielding a simplified distinguishability objective of the form~$\sum_{z\in C}  s(\theta^*, z) - s(\theta_C^*, z)$.

This objective is meaningful only when interference between canaries remains sufficiently low. \Cref{alg:canary_selection} enforces this property at initialization, but subsequent gradient updates may reintroduce interference as the inner model and canaries evolve. We therefore introduce a regularizer $\calR_{\theta^*}(C)$ that preserves this structure throughout the optimization, leading to the following bilevel problem:
\begin{align}\label{eq:regularized_pb}
  \min_{C}\; \Phi(C)\triangleq \sum_{z\in C} s(\theta^*, z) - s(\theta_C^*, z) + \calR_{\theta^*}(C)
  \;\text{s.t.}\;
  \begin{cases}
    \theta^*\in\argmin_{\theta\in\Theta} \calL(\theta, D),\\
    \theta_C^*\in\argmin_{\theta\in\Theta} \calL(\theta, D\cup C).
  \end{cases}
\end{align}
Building on the intuition from \cref{ssec:warmup}, we define the regularizer in terms of representation-space orthogonality. Denoting by $C_x$ the inputs of the canaries in $C$, we set
$
  \calR_{\theta^*}(C) = \lambda\sum_{x,x'\in C_x,x\neq x'}\langle \varphi_{\theta^*}(x), \varphi_{\theta^*}(x')\rangle^2,
$
where $\varphi_{\theta^*}:\mathbb{R}^d \to \mathbb{R}^{d'}$ maps each input to its embedding under $\theta^*$, and $\lambda > 0$ trades off between reducing mutual interference and preserving canary detectability. Using $\theta^*$ rather than $\theta^*_C$ decouples the regularizer from the inner optimization problem, which avoids additional complexity.
In the bilevel problem \eqref{eq:regularized_pb}, only $\theta_C^*$ depends on the canary set. The model $\theta^*$ can therefore be computed once in advance, while $\theta_C^*$ is updated incrementally as $C$ evolves rather than recomputed from scratch, enabling amortized bilevel schemes \citep{Ji2021stocbio,Arbel2022amigo}.\looseness=-1

\subsection{Efficient gradient-based canary crafting}

We now introduce an efficient algorithm to solve the bilevel optimization problem defined in the previous section.
The optimization is initialized using the canary set obtained from \cref{sec:preselection}, and then refined via gradient-based updates.
To avoid discrete optimization over labels, we keep them fixed throughout training and optimize only the canary inputs $C_x$.

To perform gradient-based optimization of $\Phi$, we require its gradient with respect to $C_x$, denoted by $\nabla_{C_x} \Phi$, which is referred to as the hypergradient in the bilevel optimization literature. \Cref{prop:hypergradient} provides an explicit expression for $\nabla_{C_x} \Phi$ for the case where the inner problem is strongly convex. For a canary $z = (x,y)$, the notation $\nabla_x$ indicates that differentiation is performed with respect to the input alone, the label being held fixed.

\begin{proposition}\label{prop:hypergradient}
    Assume that the score function $s$ is differentiable w.r.t. both $\theta$ and the input sample $x$, and that the loss function $\ell$ is twice differentiable, strongly convex in $\theta$ and differentiable in $x$. Then $\Phi$ is differentiable and, letting $n \triangleq \lvert D\rvert+\lvert C\rvert$, its gradient is given by
    \begin{align*}
        \nabla_{C_x} \Phi(C) = \sum_{z\in C}\big[\nabla_x s(\theta^*, z) - \nabla_x s(\theta_C^*, z) - \tfrac{1}{n}\nabla_{x,\theta}^2\ell(\theta_C^*, z)v^*(C)\big] + \nabla_{C_x}\calR_{\theta^*}(C)
    \end{align*}
    where $v^*(C)$ is the solution of the linear system $\nabla^2_{\theta,\theta}\calL(\theta_C^*, D\cup C)v = - \sum_{z\in C}\nabla_\theta s(\theta_C^*, z)$.
\end{proposition}
 
Proposition~\ref{prop:hypergradient} follows by applying the standard implicit differentiation result for bilevel optimization to our objective; we provide the proof in \cref{app:proof_hypergradient}. While \cref{prop:hypergradient} provides an exact expression, directly using it to perform gradient descent is intractable in large-scale scenarios. Indeed, it requires repeatedly recomputing $\theta_C^*$ as the canary set evolves, effectively retraining the model at each update, as well as solving a high-dimensional linear system to obtain $v^*(C)$. To address this, we rely on Approximate Implicit Differentiation (AID) \citep{Pedregosa2016hoag,Ghadimi2018}, which replaces $\theta_C^*$ and $v^*(C)$ with estimates updated during optimization. Although AID is theoretically grounded in settings with strongly convex inner problems, it has also been successfully applied in non-convex regimes~\citep{Grangier2024}.

Concretely, we instantiate this approach with the SOBA algorithm, which converges to a stationary point of $\Phi$ when the inner problem is strongly convex \citep[Theorems 1 and 2]{Dagreou2022SABA}. The algorithm maintains three coupled variables $\theta$, $v$ and $C_x$. At iteration $t$, the iterates $\theta_t$ and $v_t$ approximate $\theta^*(D\cup C_t)$ and $v^*(C_t)$ respectively. They are updated using stochastic gradient steps to track their respective equilibria as the canary set evolves, while $C_x$ is updated via an approximate hypergradient step. The resulting recursion is:
\begin{align}\label{eq:soba_update_theta}
    \theta_{t+1} &= \theta_t - \eta \nabla_\theta\calL(\theta_t, B_t)\\\label{eq:soba_update_v}
    v_{t+1} &= \textstyle v_t - \eta\Big[\nabla^2_{\theta,\theta}\calL(\theta_t, B_t)v_t + \sum_{z\in C_t}\nabla_\theta s(\theta_t, z)\Big]\\\label{eq:soba_update_x}
    C_{x,t+1} &= \textstyle  C_{x,t} - \rho \Big[\sum_{z\in C_t}\big(\nabla_x s(\theta^*, z) - \nabla_x s(\theta_t, z) - \tfrac{1}{n}\nabla^2_{x,\theta}\ell(\theta_t, z)v_t\big) + \nabla_{C_x}\calR_{\theta^*}(C_t)\Big]
\end{align}
where $B_t\subset D\cup C_t$ is a mini-batch of samples. Note that \eqref{eq:soba_update_x} coincides with the exact hypergradient update when $\theta_t=\theta_{C_t}^*$ and $v_t = v^*(C_t)$. Moreover, although the recursion involves Hessian-vector products, these can be computed efficiently without explicitly forming the Hessian using the Pearlmutter trick \citep{Pearlmutter1994,Dagreou2024hvp}.
Thanks to the structure of our bilevel formulation, which compares the model trained with all canaries against a fixed reference model, \eqref{eq:soba_update_theta}--\eqref{eq:soba_update_x} avoid full retraining at each canary update. Instead, the model parameters $\theta$ evolve jointly with the canary optimization. In practice, we initialize $\theta^0$ and $v^0$ by approximately computing $\theta_{C_0}^*$ and $v^*(C_0)$, ensuring that for small enough step sizes, we keep $\theta_t\approx \theta_{C_t}^*$ and $v_t\approx v^*(C_t)$ throughout optimization.
The recursion also supports per-example gradient clipping and data augmentation (see \cref{app:exp_adapt}), allowing our method to account for these components when auditing models that use them.\looseness=-1

\subsection{IBIS — Influence-Bilevel canary crafting with Interference Suppression}

\begin{algorithm}[t]
  \begin{algorithmic}[1]
    \STATE {\bfseries Input:} Dataset $D$, score $s$, $m$, $p$, $\lambda$, $\eta$, $\rho$, $T$
    \STATE Get $C_0$ via \cref{alg:canary_selection}; set $D\leftarrow D\setminus C_0$.
    \STATE Compute $\theta^*\approx\argmin_\theta\calL(\theta,D)$; initialize $\theta_0\approx\theta^*(D\cup C_0)$, $v_0\approx v^*(C_0)$.
    \FOR{$t=0,\dots,T-1$}
      \STATE Sample $B_t\subset D\cup C_t$;
      \STATE Update $\theta_{t+1}$ via \eqref{eq:soba_update_theta}.
      \STATE Update $v_{t+1}$ via \eqref{eq:soba_update_v}.
      \STATE Update $C_{x,t+1}$ via \eqref{eq:soba_update_x}.
    \ENDFOR
    \STATE {\bfseries Output:} $C_T$.
  \end{algorithmic}
  \caption{\textbf{IBIS}: Influence-Bilevel canary crafting with Interference Suppression}
  \label{alg:full_pipeline}
\end{algorithm}
Our full procedure, which we call \textbf{IBIS}, is summarized in \cref{alg:full_pipeline}. It combines the two contributions of this paper into a unified pipeline: the influence-based greedy selection (\cref{alg:canary_selection}) provides an initialization with strong detectability and low pairwise interference, and the bilevel updates \eqref{eq:soba_update_theta}--\eqref{eq:soba_update_x} further refine the canaries while preserving their diversity through the regularizer $\calR_{\theta^*}$.
Overall, the procedure requires three model trainings: one for influence-based selection and two to initialize $\theta^*$ and $\theta_0\approx\theta^*(D\cup C_0)$ before the bilevel loop. This modest overhead relative to the cost of training the audited model preserves the computational advantage of one-run over multi-run auditing.\looseness=-1

\textbf{Difference with \citet{Boglioni2026meta_learn_canaries}.} The canary crafting algorithm of \citet{Boglioni2026meta_learn_canaries} can also be formulated as a bilevel optimization problem. It learns canary sets through gradient steps obtained by differentiating through SGD updates, which approximately solves the following problem by iterative differentiation~\citep{Domke2012,Maclaurin2015,Bolte2023one_step_diff}:\looseness=-1
\begin{align}\label{eq:pb_boglioni}
\min_{C} \Psi(C) \triangleq \bbE_S \bigg[\sum_{z\in C\setminus S}s(\theta_S^*, z) - \sum_{z\in S}s(\theta_S^*, z)\bigg] \;\text{s.t.}\; \theta_S^*\in\argmin_{\theta\in\Theta} \calL(\theta, D\cup S),\;\forall S\subset C,
\end{align}
where the expectation is over the random subset $S$, formed by including each canary independently with probability $1/2$. Minimizing $\Psi$ encourages included canaries to have high scores and excluded canaries to have low scores, averaged over subsets $S$.
Unlike \eqref{eq:regularized_pb}, this formulation requires a distinct inner solution $\theta_S^*$ for every subset $S\subset C$. Consequently, whereas our approach performs a single gradient update as in \eqref{eq:soba_update_theta}, their approach must retrain the model for each sampled $S$. This computation cannot be amortized without maintaining a separate model for every possible subset. We compare the wall-clock time of the two approaches in \cref{ssec:boglioni}, where our approach is two orders of magnitude faster. %

\section{Experiments}
We audit WRN16-4 models~\citep{Zagoruyko2016wrn} trained on CIFAR10 and CIFAR100, and a multilayer perceptron (MLP) trained on MNIST. Following \citet[Appendix E.8]{Yaghini2026optifluence}, we train the WRN16-4 models with per-epoch data augmentation and an exponential moving average of the weights to avoid blatant memorization, as shown in \cref{app:no_aug}.
Further experimental details are given in \cref{app:exp_details}.

\subsection{Role of initialization and regularization in IBIS}

We study the role of initialization and regularization in IBIS from two complementary perspectives: first, through an ablation study, and second, by directly evaluating whether our orthogonality-based regularization reduces interference and thereby improves audit performance.

\begin{figure}[h]
  \centering
  \begin{subfigure}{0.49\linewidth}
    \centering
    \includegraphics[width=\linewidth]{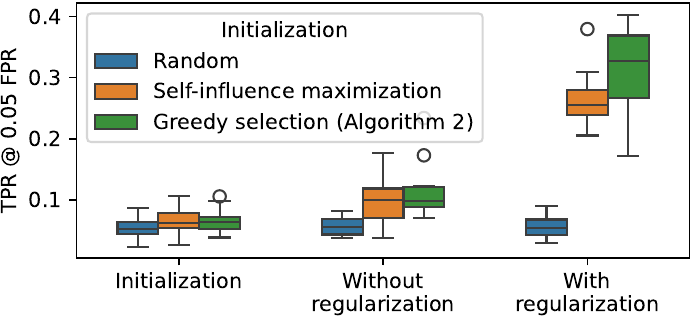}
    \caption{Impact of initialization and regularization on the audit performance (TPR@0.05FPR).}
    \label{fig:ablation}
  \end{subfigure}
  \hfill
  \begin{subfigure}{0.49\linewidth}
    \centering
    \includegraphics[width=\linewidth]{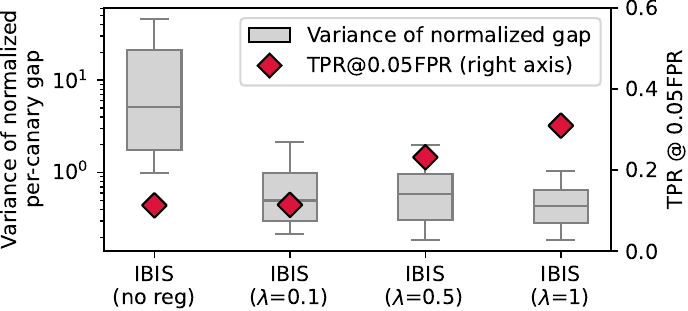}
    \caption{Estimated interference and audit performance (TPR@0.05FPR) for several regularization strengths $\lambda$.}
    \label{fig:interference_per_canary}
  \end{subfigure}
  \caption{Role of initialization and regularization in IBIS (WRN16-4 on CIFAR10).}
  \label{fig:ablation_interference}
\end{figure}

\paragraph{Effectiveness of initialization and regularization.} We conduct an ablation study on WRN16-4 trained on CIFAR10. We initialize the bilevel algorithm with $m=1000$ canaries selected either randomly, by maximizing self-influence, or using our proposed greedy selection strategy (\cref{alg:canary_selection}). For each initialization strategy, we run IBIS with and without regularization ($\lambda=1$ and $\lambda=0$, respectively) and report the resulting TPR@0.05FPR in \Cref{fig:ablation}. With influence-based initialization, the unregularized bilevel optimization yields only a moderate improvement, whereas regularization leads to a substantial performance gain. The magnitude of this gain depends on the initialization strategy: with regularization, our proposed greedy selection achieves higher scores than self-influence maximization.
These results highlight the importance of influence-based initialization and the value of accounting for cross-influence when selecting canaries for one-run privacy auditing.

\paragraph{Interference and audit performance.} Our regularization uses orthogonality in embedding space as a proxy for reducing interference. We thus vary the regularization strength $\lambda$ and measure its effect on both interference and downstream audit performance. For a canary $z\in C$ and a subset $S\subset C$, we define the score gap $\Delta(z, S) \triangleq s(\calA(D\cup S \cup\{z\}), z) - s(\calA(D\cup S \setminus\{z\}), z)$. In the absence of interference, this gap is independent of $S$. We therefore quantify interference by $\Var_S[\Delta(z, S)]$, normalized by the squared mean gap across canaries. As shown in \Cref{fig:interference_per_canary} regularization reduces this measure by about an order of magnitude, while increasing TPR@0.05FPR from 0.11 to 0.31.

\subsection{Comparison with Standard Canary Constructions} \label{sec:evaluation}

We compare IBIS with two standard canary constructions: Random and Mislabeled. The comparison with the method of \citet{Boglioni2026meta_learn_canaries} is deferred to \cref{ssec:boglioni}. Random canaries are images drawn uniformly at random from the dataset. Mislabeled canaries are likewise drawn at random, but their labels are flipped.

\Cref{tab:mia_wrn_cifar10} reports audit performance in terms of TPR@0.05FPR. Since an auditor typically performs a single audit, we also report the minimum value across runs, which captures the leakage that can be reliably detected. Random canaries remain close to the 0.05 level expected under random guessing, indicating that typical training samples are difficult to detect in a single run. On CIFAR10, IBIS canaries achieve higher mean and minimum TPR than Mislabeled canaries. On CIFAR100, IBIS achieves slightly higher scores on both metrics.

\begin{table}[t]
  \caption{Audit performance (TPR@0.05FPR) for a WRN16-4 model on CIFAR10 and CIFAR100. We report mean, standard deviation and minimum over 15 runs.}
  \label{tab:mia_wrn_cifar10}
  \centering
  \begin{tabular}{lcccc}
    \toprule
    & \multicolumn{2}{c}{CIFAR10} & \multicolumn{2}{c}{CIFAR100} \\
    \cmidrule(lr){2-3}\cmidrule(lr){4-5}
    & Mean $\pm$ std & Min & Mean $\pm$ std & Min \\
    \midrule
    IBIS       & $\boldsymbol{0.31 \pm 0.07}$ & \textbf{0.17} & $\boldsymbol{0.35 \pm 0.05}$ & \textbf{0.27} \\
    Random     & $0.05 \pm 0.01$ & 0.02 & $0.06 \pm 0.01$ & 0.04 \\
    Mislabeled & $0.22 \pm 0.06$ & 0.12 & $0.33 \pm 0.04$ & 0.24 \\
    \bottomrule
  \end{tabular}
\end{table}

\subsection{Comparison with \textsc{Metagradient} \citep{Boglioni2026meta_learn_canaries}}\label{ssec:boglioni}

We compare IBIS with its closest competitor, the method of \citet{Boglioni2026meta_learn_canaries}, in terms of both audit performance and computational cost. We refer to this method as \textsc{Metagradient}. For IBIS, crafting time includes both the greedy initialization and the bilevel optimization stage. On CIFAR10, we use the \textsc{Metagradient} canaries released with the authors' code, as reproducing them would be prohibitively expensive. On MNIST, no pre-crafted canaries are available, so we craft them using the authors' implementation. We do not evaluate \textsc{Metagradient} on CIFAR100, as crafting canaries with this method would require prohibitively large computational resources. \cref{tab:boglioni} reports both TPR@0.05FPR and the AUC, as \textsc{Metagradient}'s TPR@0.05FPR is close to random guessing on MNIST.
IBIS achieves higher audit performance in both metrics, while requiring up to two orders of magnitude less computation than \textsc{Metagradient}.

\begin{table}[t]
  \caption{Audit performance (AUC, TPR@0.05FPR) and canary crafting runtime of IBIS and \textsc{Metagradient} on CIFAR10 and MNIST. We report mean and standard deviation over 15 runs. Times marked with * are those reported by \citet[Appendix B]{Boglioni2026meta_learn_canaries}.}
  \label{tab:boglioni}
  \centering
  \small
  \setlength{\tabcolsep}{4pt}
  \begin{tabular}{lcccccc}
    \toprule
    & \multicolumn{3}{c}{IBIS} & \multicolumn{3}{c}{\textsc{Metagradient} \citep{Boglioni2026meta_learn_canaries}} \\
    \cmidrule(lr){2-4}\cmidrule(lr){5-7}
    & AUC & TPR@0.05FPR & Time [h] & AUC & TPR@0.05FPR & Time [h] \\
    \midrule
    CIFAR10 & $\boldsymbol{0.82 \pm 0.02}$ & $\boldsymbol{0.31 \pm 0.07}$ & \textbf{1.2} & $0.76 \pm 0.03$ & $0.15 \pm 0.05$ & 96--120* \\
    MNIST   & $\boldsymbol{0.86 \pm 0.02}$ & $\boldsymbol{0.29 \pm 0.07}$ & \textbf{0.1} & $0.75 \pm 0.01$ & $0.06 \pm 0.01$ & 30 \\
    \bottomrule
  \end{tabular}
\end{table}

\textbf{Additional results.} In \cref{app:exp_dp} we provide an experiment of DP auditing. In this setting, IBIS and \textsc{Metagradient} canaries achieve comparable results. In \cref{app:exp_multi_run}, we compare the whole pipeline of multi-run and one-run auditing, showing that crafting canaries with IBIS does not cancel the computational advantage of one-run auditing.

\section{Conclusion}
In this work, we studied canary crafting for one-run privacy auditing. Building on the observation that interference between canaries can limit the power of such audits, we propose IBIS, which uses influence functions to select detectable, weakly interfering canaries and a bilevel objective to further reduce interference by promoting orthogonality in embedding space.  This low-interference regime allows IBIS to update a single model trained with all canaries incrementally, avoiding retraining at every canary update. Experiments show that IBIS improves audit performance over existing canary crafting methods while reducing crafting cost by up to two orders of magnitude.
While one-run auditing substantially reduces computational cost, its audit performance remains lower than that of multi-run auditing, reflecting the challenges of black-box auditing from a single model. Whether improved canary crafting can close this gap in some settings remains an open question.

\subsection*{AI use statement}

In this work, large language models were used to polish the writing and format the tables of the paper. The LLM-generated content was reviewed by the authors, who take full responsibility for the final content of this work.

\subsection*{Reproducibility statement}

Proofs of the theoretical results are given in \cref{app:proof_gap_canary_ols,app:proof_hypergradient}. The experimental details, including all hyperparameters, are given in \cref{app:exp_details}. The anonymized source code is provided as supplementary material.

\subsubsection*{Acknowledgments}
This work is partially supported by grant ANR-22-PESN-0014 project of the Digital Health PEPR under the France 2030 program. AB is partially supported by ANR 22-PECY-0002 IPOP (Interdisciplinary Project on Privacy) project of the Cybersecurity PEPR. NP is partly supported by an Inria International Chair. Part of this work was performed using HPC resources from GENCI–IDRIS (Grant 2025-AD011016328). Part of the experiments presented were carried out using the Grid'5000 testbed, supported by a scientific interest group hosted by Inria and including CNRS, RENATER and several Universities as well as other organizations (see \url{https://www.grid5000.fr}).

\bibliography{biblio}
\bibliographystyle{bibstyle}

\appendix
\section{Proofs}
\subsection{\texorpdfstring{\cref{prop:gap_canary_ols}}{Proposition 3.1}}\label{app:proof_gap_canary_ols}
For convenience, we denote $\theta^*\triangleq \theta^*(X, y)$ and $\theta^*(\tilde X_1,\tilde y_1) = \theta^*_1$. We have 
\begin{align*}
  \theta^*_1  = \big[\tilde X_1^\top \tilde X_1\big]^{-1}\tilde X_1^\top \tilde y_1
              =  \Big[\sum_{i=1}^n x_ix_i^\top + x_{c,1}x_{c,1}^\top\Big]^{-1}\Big[\sum_{i=1}^ny_ix_i + y_{c,1}x_{c,1}\Big]
\end{align*}
Let us denote $K = X^\top X$. By the Sherman-Morrison formula \cite{Sherman1950}, we have
\begin{align*}
  \theta^*_1  &= \big[\tilde X_1^\top \tilde X_1\big]^{-1}\tilde X_1^\top \tilde y_1 \\
              &= \Big(K^{-1} - \frac{K^{-1}x_{c,1}x_{c,1}^\top K^{-1}}{1+x_{c,1}^\top K^{-1}x_{c,1}}\Big)\tilde X_1^\top \tilde y_1 \\
              &= \Big(I-\frac{K^{-1}x_{c,1}x_{c,1}^\top}{1+x_{c,1}^\top K^{-1}x_{c,1}}\Big)K^{-1}\tilde X_1^\top \tilde y_1 \\
              &= \Big(I-\frac{K^{-1}x_{c,1}x_{c,1}^\top}{1+x_{c,1}^\top K^{-1}x_{c,1}}\Big)(\theta^* + y_{c,1}K^{-1}x_{c,1}) \\
              &= \theta^*  + y_{c,1}K^{-1}x_{c,1} - \frac{K^{-1}x_{c,1}x_{c,1}^\top}{1+x_{c,1}^\top K^{-1}x_{c,1}}(K^{-1}y_{c,1} x_{c,1} + \theta^*)\\
              &= \theta^*  + \left(y_{c,1}-\frac{y_{c,1}x_{c,1}^\top K^{-1}x_{c,1} + \langle\theta^*,x_{c,1}\rangle }{1+x_{c,1}^\top K^{-1}x_{c,1}}\right)K^{-1}x_{c,1}\\
              &=  \theta^*  -\frac{K^{-1}}{1+x_{c,1}^\top K^{-1} x_{c,1}}(\langle \theta^*, x_{c,1}\rangle - y_{c,1})x_{c,1}.
\end{align*}

Therefore we have
\begin{align*}
    \ell(\theta^*_1,z_{c,2} ) &= \frac12\left(\langle \theta^*_1, x_{c,2}\rangle - y_{c,2}\right)^2\\
    &= \ell(\theta^*, z_{c,2}) + \ell(\theta^*, z_{c,1})\left(\frac{x_{c,1}^\top K^{-1}x_{c,2}}{1 + x_{c,1}^\top K^{-1} x_{c,1}}\right)^2 \\
    &\quad - (\langle \theta^*, x_{c,1}\rangle -  y_{c,1})(\langle \theta^*, x_{c,2}\rangle -  y_{c,2})\frac{x_{c,1}^\top K^{-1}x_{c,2}}{1 + x_{c,1}^\top K^{-1} x_{c,1}}.
\end{align*}
Both remaining terms carry the factor $x_{c,1}^\top K^{-1}x_{c,2}$. Denoting by $r_i = \langle\theta^*, x_{c,i}\rangle - y_{c,i}$ the residual at the canary $z_{c,i}$ and factoring this term out gives $  \ell(\theta^*_1, z_{c,2}) - \ell(\theta^*, z_{c,2}) = \gamma\; x_{c,1}^\top K^{-1}x_{c,2}$ with
\begin{align*}
  \gamma = \frac{\ell(\theta^*, z_{c,1})\,x_{c,1}^\top K^{-1}x_{c,2}}{\big(1 + x_{c,1}^\top K^{-1}x_{c,1}\big)^2} - \frac{r_1 r_2}{1 + x_{c,1}^\top K^{-1}x_{c,1}},
\end{align*}
which yields the result.

\subsection{\texorpdfstring{\cref{prop:hypergradient}}{Proposition 4.1}}\label{app:proof_hypergradient}
Recall that the function $\Phi$ is defined as 
$$
    \Phi(C)= \sum_{z\in C} s(\theta^*, z) - s(\theta_C^*, z) + \calR_{\theta^*}(C)
$$
with $\theta^*\in\argmin_{\theta\in\Theta}\calL(\theta, D)$ and $\theta^*_C\in\argmin_{\theta\in\Theta}\calL(\theta, D\cup C)$. By the chain rule, we get:
$$
  \nabla_{C_x} \Phi(C)= \sum_{z\in C} \Big[\nabla_x s(\theta^*, z) - \nabla_x s(\theta_C^*, z) - (\diff \theta^*_C)^\top \nabla_\theta s(\theta_C^*, z)\Big] + \nabla_{C_x}\calR_{\theta^*}(C)
$$
where $\diff \theta^*_C$ is the Jacobian of $(C_x\mapsto \theta^*_C)$. 
Since $\calL(\,\cdot\,,D)$ and $\calL(\,\cdot\,,D\cup C)$ are strongly convex, the value function $\Phi$ is well-defined. Moreover, for any canary sample $C$, $\theta^*_C$ verifies the following first order condition
\begin{equation}\label{eq:stationarity}
    \nabla_\theta\calL(\theta^*_C, D\cup C) = 0.
\end{equation}
The Hessian matrix $\nabla_{\theta,\theta}^2\calL(\theta^*_C, D\cup C)$ is positive definite by strong convexity and since $\ell$ is differentiable with respect to the input samples $C_x$, by the Implicit Function Theorem, we can differentiate Equation \eqref{eq:stationarity} with respect to $C_x$ to get 
$$
  \nabla^2_{\theta,\theta}\calL(\theta^*_C, D\cup C)\diff \theta^*_C + \nabla^2_{C_x,\theta}\calL(\theta^*_C, D\cup C) = 0.
$$
Rearranging yields the result.

\section{Additional experiments}

\subsection{Adapting the updates to the audited training pipeline}\label{app:exp_adapt}

The update rules \eqref{eq:soba_update_theta}--\eqref{eq:soba_update_x} assume that the audited model minimizes the empirical risk $\calL(\theta, D\cup C)$. Since the inner problem of \eqref{eq:regularized_pb} models the audited training, we can adapt it to common components of the training pipeline, e.g. data augmentation and per-sample gradient clipping.

\paragraph{Data augmentation.} When the audited model is trained with data augmentation, we apply the same augmentation to each mini-batch $B_t$ in \eqref{eq:soba_update_theta} and \eqref{eq:soba_update_v}. \Cref{tab:adaptation} shows that applying the augmentation during the crafting increases the TPR@0.05FPR from 0.21 to 0.31 when the audited model is trained with data augmentation.

\paragraph{Per-example gradient clipping.} DP-SGD clips each per-example gradient to a norm of at most $C$ before averaging. Without noise, a model trained this way does not converge to a minimizer of $\calL(\theta, D\cup C)$, but to a point $\theta^*_C$ that satisfies
$$G(\theta^*_C, D\cup C) \triangleq \frac{1}{n}\sum_{z\in D\cup C}\clip_C\big[\nabla_\theta \ell(\theta^*_C, z)\big] = 0,$$
where $\clip_C(g) = g\min(1, C/\|g\|)$. Applying the implicit function theorem to this equation yields the hypergradient of \cref{prop:hypergradient}, where the Hessian $\nabla^2_{\theta,\theta}\calL$ is replaced by the Jacobian $\nabla_\theta G$, and the cross-derivative $\nabla^2_{x,\theta}\ell$ by the derivative of the clipped gradient with respect to the canary input. We modify \eqref{eq:soba_update_theta}--\eqref{eq:soba_update_x} accordingly. The iterate $\theta_t$ follows the clipped gradient, and the Jacobian-vector and cross-derivative products are computed through the clipping operator by automatic differentiation. \Cref{tab:adaptation} shows that accounting for clipping improves both the lower bound $\hat\epsilon$ and the TPR@0.05FPR when auditing DP-SGD at $\epsilon = 10$.

\begin{table}[h]
  \caption{Effect of adapting IBIS to the audited training pipeline, on WRN16-4 trained on CIFAR10. \textbf{Top:} non-private training with data augmentation and EMA, where the adapted IBIS applies the augmentation during crafting. \textbf{Bottom:} DP-SGD training at $\epsilon = 10$ and $\delta = 10^{-5}$, where the adapted IBIS incorporates clipping. We report mean and standard deviation over 15 runs.}
  \label{tab:adaptation}
  \centering
  \begin{tabular}{lcc}
    \toprule
    & IBIS & Adapted IBIS \\
    \midrule
    \multicolumn{3}{l}{\textit{Data augmentation, non-private}} \\
    TPR@0.05FPR          & $0.21 \pm 0.06$   & $\boldsymbol{0.31 \pm 0.07}$ \\
    \midrule
    \multicolumn{3}{l}{\textit{Clipping, DP-SGD at $\epsilon = 10$}} \\
    Lower bound $\hat\epsilon$ & $0.72 \pm 0.41$   & $\boldsymbol{0.98 \pm 0.38}$ \\
    TPR@0.05FPR          & $0.074 \pm 0.017$ & $\boldsymbol{0.094 \pm 0.024}$ \\
    \bottomrule
  \end{tabular}
\end{table}

\subsection{Auditing differential privacy in one run}\label{app:exp_dp}

In this subsection, we use our canaries to perform one-run privacy auditing. In this experiment, we consider the WRN16-4 architecture trained on CIFAR10 with DP-SGD, and without the exponential moving average and data augmentation used previously. Accordingly, we craft IBIS canaries with the clipping-aware variant described in \cref{app:exp_adapt}. We audit models trained with $\epsilon\in\{1, 2, 4, 10\}$ and $\delta = 10^{-5}$ and compute an empirical lower bound $\hat\epsilon$ using the GDP-based one-run auditing procedure \citep{Mahloujifar2025} since it is known to provide tighter results than the original one-run technique \citep{Steinke2023}. For each method, the same canaries are used for all values of $\epsilon$.\looseness=-1

For both metrics and at all privacy levels, the gap between IBIS and \textsc{Metagradient} canaries \citep{Boglioni2026meta_learn_canaries} remains within one standard deviation. At $\epsilon = 10$, both yield clearly higher scores than random and mislabeled canaries. At $\epsilon = 1$, all methods yield lower bounds close to zero. Overall, the lower bounds exhibit high variance across runs, which reflects the difficulty of one-run auditing under DP. IBIS thus reaches an audit quality comparable to \textsc{Metagradient} canaries at a fraction of their crafting cost, as shown in \cref{ssec:boglioni}.

\begin{table}[h]
  \caption{Empirical lower bound $\hat\epsilon$ and TPR@0.05FPR obtained by one-run auditing of WRN16-4 trained on CIFAR10 with DP-SGD ($\delta = 10^{-5}$). We report mean and standard deviation over 15 runs.}
  \label{tab:audit_gdp_wrn_cifar10}
  \centering
  \begin{tabular}{lcccc}
    \toprule
    $\epsilon$ & IBIS & \textsc{Metagradient} & Random & Mislabeled \\
    \midrule
    \multicolumn{5}{l}{\textit{Lower bound $\hat\epsilon$}} \\
    10 & $0.98 \pm 0.38$ & $0.86 \pm 0.30$ & $0.21 \pm 0.20$ & $0.45 \pm 0.43$ \\
    4  & $0.21 \pm 0.21$ & $0.38 \pm 0.32$ & $0.25 \pm 0.32$ & $0.16 \pm 0.16$ \\
    2  & $0.18 \pm 0.26$ & $0.32 \pm 0.31$ & $0.15 \pm 0.17$ & $0.11 \pm 0.10$ \\
    1  & $0.13 \pm 0.17$ & $0.12 \pm 0.13$ & $0.08 \pm 0.14$ & $0.09 \pm 0.11$ \\
    \midrule
    \multicolumn{5}{l}{\textit{TPR@0.05FPR}} \\
    10 & $0.094 \pm 0.024$ & $0.088 \pm 0.023$ & $0.047 \pm 0.016$ & $0.065 \pm 0.020$ \\
    4  & $0.063 \pm 0.013$ & $0.068 \pm 0.013$ & $0.050 \pm 0.018$ & $0.056 \pm 0.015$ \\
    2  & $0.057 \pm 0.017$ & $0.061 \pm 0.015$ & $0.051 \pm 0.015$ & $0.049 \pm 0.015$ \\
    1  & $0.055 \pm 0.013$ & $0.056 \pm 0.008$ & $0.052 \pm 0.017$ & $0.054 \pm 0.012$ \\
    \bottomrule
  \end{tabular}
\end{table}

\subsection{Comparison between multi-run and one-run auditing}\label{app:exp_multi_run}
We compare the whole pipeline of multi-run and one-run auditing on WRN16-4 trained on CIFAR10 with data augmentation and EMA. For multi-run auditing, we craft a single canary with \textsc{Optifluence} \citep{Yaghini2026optifluence}, which is the multi-run counterpart of IBIS, and audit it over 1288 training runs. For one-run auditing, we craft 1000 canaries with IBIS and audit them with a single training run. \Cref{tab:multi_vs_one_run} reports the crafting time, the audit time and the TPR@0.05FPR. The two TPR@0.05FPR are not computed on the same statistics. In multi-run auditing, it is computed over training runs for a single canary. In one-run auditing, it is computed over canaries within a single run. Multi-run auditing yields a higher TPR@0.05FPR, but its audit requires more than 160 GPU hours. With IBIS, the whole one-run pipeline takes about 1.3 GPU hours, which is two orders of magnitude cheaper. Crafting canaries with IBIS therefore preserves the computational advantage of one-run auditing.

\begin{table}[h]
  \caption{Crafting time, audit time and TPR@0.05FPR of multi-run and one-run auditing of WRN16-4 trained on CIFAR10 with data augmentation and EMA. Times are in GPU hours on an NVIDIA A100. The crafting time of \textsc{Optifluence} is measured in the setting without data augmentation, which uses the same crafting procedure. The one-run TPR@0.05FPR is averaged over 15 runs.}
  \label{tab:multi_vs_one_run}
  \centering
  \begin{tabular}{llccc}
    \toprule
    & Canary & Crafting [h] & Audit [h] & TPR@0.05FPR \\
    \midrule
    Multi-run & \textsc{Optifluence} & 2.5 & 167.5 & 0.72 \\
    One-run   & IBIS                 & 1.2 & 0.13  & 0.31 \\
    \bottomrule
  \end{tabular}
\end{table}

\subsection{Necessity of data augmentation and exponential moving average}\label{app:no_aug}

Following \citet{Yaghini2026optifluence}, we chose to incorporate data augmentation and an exponential moving average of the weights to avoid saturating the audit metrics. Indeed, when WRN16-4 is trained on CIFAR10 without these components, the IBIS and Mislabeled canaries saturate the TPR@0.05FPR, as shown in \cref{tab:no_aug}.

\begin{table}[h]
  \caption{TPR@0.05FPR of one-run auditing of WRN16-4 trained on CIFAR10 without data augmentation and EMA. We report mean and standard deviation over 15 runs.}
  \label{tab:no_aug}
  \centering
  \begin{tabular}{lccc}
    \toprule
    & IBIS & Random & Mislabeled \\
    \midrule
    TPR@0.05FPR & $1.00 \pm 0.00$ & $0.06 \pm 0.01$ & $1.00 \pm 0.00$ \\
    \bottomrule
  \end{tabular}
\end{table}

\section{Experimental details}\label{app:exp_details}
\subsection{Setting}
The experiments are run in the \texttt{PyTorch} framework \cite{Paszke2019}. For DP-SGD, we use the \texttt{opacus} implementation \cite{Yousefpour2021opacus}. For the computation of the influence scores, we use the \texttt{kronfluence} package~\cite{Grosse2023}. Unless stated otherwise, we craft 5 canary sets with different random seeds and audit each of them with 3 independent training runs. We report mean and standard deviation over these 15 runs.

\subsection{Hyperparameters}

We report in \cref{tab:hp_train} the hyperparameters used for model training and in \cref{tab:hp_canary} those used for the canary optimization pipeline. The inner learning rate $\eta$, the outer learning rate $\rho$ and the regularization parameter $\lambda$ are selected by grid search. For IBIS, the hyperparameters of the ablation study are given in \cref{tab:hp_ablation}. In \cref{fig:interference_per_canary}, we draw 20 canaries at random from each canary set. For each of them, the variance $\Var_S[\Delta(z, S)]$ is estimated from 10 random subsets $S$.

\begin{table}[h]
  \caption{Training hyperparameters of WRN16-4. The same values are used on CIFAR10 and CIFAR100. For DP-SGD, the noise multiplier is calibrated with \texttt{opacus} to reach the target $(\epsilon, \delta)$.}
  \label{tab:hp_train}
  \centering
  \small
  \begin{tabular}{lccc}
    \toprule
    & \multicolumn{2}{c}{Non-private} & \\
    \cmidrule(lr){2-3}
    Hyperparameter & Without aug. & With aug.\ and EMA & DP-SGD \\
    \midrule
    Optimizer          & SGD       & SGD       & SGD \\
    Learning rate      & 0.1       & 0.1       & see below \\
    Momentum           & 0.9       & 0.9       & 0 \\
    Weight decay       & $10^{-5}$ & $10^{-5}$ & 0 \\
    Batch size         & 256       & 256       & 2048 \\
    Epochs             & 50        & 20        & see below \\
    Data augmentation  & --        & Random crop, horizontal flip & -- \\
    EMA decay          & --        & 0.9       & -- \\
    Clipping norm $C$  & --        & --        & 1 \\
    $\delta$           & --        & --        & $10^{-5}$ \\
    \bottomrule
  \end{tabular}

  \vspace{0.5em}
  \begin{tabular}{lcccc}
    \toprule
    DP-SGD, $\epsilon$ & 1 & 2 & 4 & 10 \\
    \midrule
    Learning rate & 1 & 1 & 2 & 2 \\
    Epochs        & 75 & 125 & 150 & 200 \\
    \bottomrule
  \end{tabular}
\end{table}

\begin{table}[h]
  \caption{Hyperparameters of IBIS (\cref{alg:full_pipeline}) on WRN16-4. The augmentation-aware and clipping-aware variants, described in \cref{app:exp_adapt}, are used to audit models trained with data augmentation and with DP-SGD, respectively.}
  \label{tab:hp_canary}
  \centering
  \small
  \begin{tabular}{lccc}
    \toprule
    & \multicolumn{2}{c}{Augmentation-aware} & Clipping-aware \\
    \cmidrule(lr){2-3}\cmidrule(lr){4-4}
    Hyperparameter & CIFAR10 & CIFAR100 & CIFAR10 \\
    \midrule
    \multicolumn{4}{l}{\textit{Influence-based selection}} \\
    Num.\ canaries $m$           & 1000  & 1000  & 1000 \\
    Preselection size $p$        & 10000 & 10000 & 10000 \\
    Batch size                   & 256   & 256   & 256 \\
    \midrule
    \multicolumn{4}{l}{\textit{SOBA}} \\
    Inner learning rate $\eta$   & 0.05  & 0.05  & 0.05 \\
    Outer learning rate $\rho$   & 1     & 1     & 100 \\
    Regularization $\lambda$     & 1     & 0.1   & 100 \\
    Epochs                       & 30    & 30    & 30 \\
    Batch size                   & 256   & 256   & 256 \\
    Clipping norm $C$            & --    & --    & 1 \\
    \bottomrule
  \end{tabular}
\end{table}

\begin{table}[h]
  \caption{Hyperparameters of IBIS in the ablation study of \cref{fig:ablation_interference}, on WRN16-4 and CIFAR10. The other hyperparameters are those of the CIFAR10 column of \cref{tab:hp_canary}.}
  \label{tab:hp_ablation}
  \centering
  \small
  \begin{tabular}{lcc}
    \toprule
    & Without reg. & With reg. \\
    \midrule
    Outer learning rate $\rho$           & 100 & 1 \\
    Regularization $\lambda$, \cref{fig:ablation}                & 0   & 1 \\
    Regularization $\lambda$, \cref{fig:interference_per_canary} & 0   & $\{0.1, 0.5, 1\}$ \\
    \bottomrule
  \end{tabular}
\end{table}

\end{document}